\documentclass[10pt,journal,compsoc]{IEEEtran}
\ifCLASSOPTIONcompsoc
  \usepackage[nocompress]{cite}
\else
  \usepackage{cite}
\fi
\ifCLASSINFOpdf
\else
\fi
\usepackage{amsmath}
\usepackage{algorithmic}
\usepackage{array}
\usepackage{fixltx2e}
\usepackage{stfloats}
\usepackage{amsfonts}
\usepackage{mathtools}
\usepackage{amssymb}
\usepackage{float}
\usepackage{adjustbox}
\usepackage{soul}
\usepackage{xcolor}

\usepackage{url}

\begin{document}
%
\title{ACTNET: end-to-end learning of feature activations and multi-stream aggregation for effective instance image retrieval}
%
%
%
%

\author{Syed~Sameed~Husain,
        Eng-Jon~Ong and
        Miroslaw~Bober,~\IEEEmembership{Member,~IEEE}
        \thanks{S. Husain, E.Ong and M. Bober are with University of Surrey, Guildford, Surrey, UK.
E-mails: \{sameed.husain, e.ong, m.bober\}@surrey.ac.uk }
}

%
%

\markboth{Journal of \LaTeX\ Class Files,~Vol.~14, No.~8, August~2015}%
{Shell \MakeLowercase{\textit{et al.}}: Bare Demo of IEEEtran.cls for Computer Society Journals}
%



\IEEEtitleabstractindextext{%
\begin{abstract}
We propose a novel CNN architecture called ACTNET for robust instance image retrieval from large-scale datasets. Our key innovation is a learnable activation layer designed to improve the signal-to-noise ratio (SNR) of deep convolutional feature maps. Further, we introduce a controlled multi-stream aggregation, where complementary deep features from different convolutional layers are optimally transformed and balanced using our novel activation layers, before aggregation into a global descriptor. Importantly, the learnable parameters of our activation blocks are explicitly trained, together with the CNN parameters, in an end-to-end manner minimising triplet loss. This means that our network jointly learns the CNN filters and their optimal activation and aggregation for retrieval tasks. To our knowledge, this is the first time parametric functions have been used to control and learn optimal multi-stream aggregation. We conduct an in-depth experimental study on three non-linear activation functions: Sine-Hyperbolic, Exponential and modified Weibull, showing that while all bring significant gains the Weibull function performs best thanks to its ability to equalise strong activations. The results clearly demonstrate that our ACTNET architecture significantly enhances the discriminative power of deep features, improving significantly over the state-of-the-art retrieval results on all datasets.
\end{abstract}

\date{\today}
}

\maketitle

\IEEEdisplaynontitleabstractindextext

\IEEEpeerreviewmaketitle

\IEEEraisesectionheading{\section{Introduction}\label{sec:introduction}}
\IEEEPARstart {T}{he} last decade has seen an explosive growth in the usage of multimedia. Conventional solutions to the management of huge volumes of multimedia fell below expectations, stimulating active research in areas such as instance image retrieval (IIR) and object recognition. Visual recognition capabilities are crucial for many applications including mobile visual search, augmented reality, robotic vision, automotive navigation, mobile commerce, surveillance \& security, content management and medical imaging, and they will only grow in importance as AI systems proliferate. In IIR, the aim is to retrieve images depicting instances of a user-specified object in a large unordered collection of images. The task is challenging as the objects to be retrieved are often surrounded by background clutter or partially occluded. Additionally, variations in the object's appearance exist due to non-linear transformations such as viewpoint and illumination variations, ageing and weather changes, etc. Consequently, robust techniques that can cope with significant variability of local image measurements and numerous outliers are required. Moreover, modern systems must be scalable due to the overwhelming volumes of multimedia data.

In order to overcome these challenges, a compact mathematical representation of image content that supports fast and robust object retrieval is required. While Convolutional Neural Networks (CNN) have contributed significantly to the best performing methods in many computer vision tasks, including image classification, they are yet to fully meet the demanding user expectations in image retrieval tasks, especially on industrial-scale datasets. This is evidenced by the best prior art result reported on a extremely challenging $R$Oxford-hard+1Million dataset at 19.9\% mAP \cite{ROXF}. (which is improved by the presented design to 29.9\%).

To bridge the gap between retrieval performance and user's expectations, we develop a novel deep neural network architecture called the Activation Network (ACTNET). In our design, we focus on areas where relatively little innovation happened; specifically on a novel and learnable aggregation framework capable of combining multi-stream dense convolutional features into a powerful global descriptor. Our architecture brings very significant performance gains, which translate into world leading retrieval rates (50\% relative gain in mAP on $R$Oxford-hard+1Million) or enable significant reduction in computational complexity (5-fold). Our main contributions are:
\begin{itemize}
	\item we propose a novel trainable activation layer that improves the signal-to-noise ratio (SNR) of deep convolutional feature maps. We conduct an in-depth experimental study to illustrate the effects of applying three non-linear activation functions: Sine-Hyperbolic, Exponential and modified Weibull. The results show that these activation functions significantly enhance the discriminative power of convolutional features, leading to world-class retrieval results.
	\item we design a multi-stream CNN architecture (ACTNET) where deep features from different convolutional layers are transformed using our novel activation layers and optimally aggregated into a compact global descriptor. Crucially, the trainable parameters from multiple CNN layers including the parameters of activation and deactivation blocks are explicitly trained in an end-to-end manner minimising the triplet loss.
	\item we develop a low-latency, small-memory and low-power ACTNET model: (1) to verify that the proposed model works within a structurally different, light architecture (i.e. very shallow MobileNet CNN), (2) to confirm that it continues bringing significant performance gains, and (3) to quantify these gains. Detailed experimental results show that our low-complexity ACTNET achieves comparable performance to the state-of-the-art RMAC \cite{Gordo2017} and GEM \cite{GEM} CNNs while providing five times faster extraction speed.
	\item we perform extensive experiments on several datasets and show that ACTNET outperforms the best CNN-based methods: MAC \cite{Tolias2015}, RMAC \cite{Gordo2017} and GEM \cite{GEM}.
\end{itemize}

The ACTNET architecture generates a robust and compact global descriptor that works extremely well for instance image retrieval. At its base, it uses any CNN model (here we apply it to two diverse models: ResNext \cite{RESNEXT} and MobileNet \cite{MobileNet}) as a powerful convolutional feature extractor. These deep features are transformed and aggregated using novel activation layers, average pooling layers and a PCA+Whitening layer. ACTNET is trained in an end-to-end manner using Stochastic Gradient Descent (SGD) with a triplet loss function to jointly optimise multi-stream features and their aggregation. The proposed method achieves retrieval performances of 74.8\%, 54.6\%, 81.5\% and 63.8\% on $R$Oxf$M$, $R$Oxf$H$, $R$Par$M$ and $R$Par$H$ datasets respectively, outperforming the latest state-of-the-art methods including those based on computationally costly local descriptor indexing and spatial verification. 

The paper is organised as follows: Section \ref{sec:lit_review} presents a literature review of existing methods. Section \ref{sec:actnet} presents in detail our novel ACTNET architecture. The experimental setup and an extensive evaluation of ACTNET is presented in Section \ref{sec:experminets}. In Section \ref{sec:SOTA} we compare our results with the state-of-the-art, demonstrating significant improvement over recent global descriptors on all retrieval datasets. Finally, Section \ref{sec:concl} concludes the paper.

\section{Related Work on Image Retrieval}\label{sec:lit_review}

This section presents an overview of systems that have contributed significantly to instance image retrieval.

\subsection{Classical image retrieval}
Classical techniques for IIR involve aggregating scale-invariant hand-crafted descriptors such as SIFT \cite{Lowe04} or SURF \cite{Bay08} into a single global image descriptor for fast matching. The most popular global representations that encode distributions of local descriptors in an image are Bag of Features (BoF) \cite{BOW}, Fisher Vectors (FV) \cite{Perronnin10}, Vector of Locally Aggregated Descriptors
(VLAD) \cite{Jegou12PAMI}, Triangulation Embedding (TEmb) \cite{Jegou14} and Robust Visual Descriptor (RVD) \cite{HusainPAMI}.  

Methods based on hand-crafted descriptors offer limited retrieval performance due to poor robustness of local features detectors and descriptors, to complex image transformations such significant view point and illumination changes.  

\subsection{Deep CNN image retrieval}
Recent approaches for IIR use the deep convolutional features of a CNN, typically trained for ImageNet classification, to compute global image representations. Azizpour et al. \cite{Razavian} perform max-pooling while Babenko et al. \cite{SPOC} apply sum-pooling to the last convolutional features of VGG16 \cite{vgg16}. One step improvement is the cross-dimensional weighted sum-pooling of Kalantidis et al. \cite{Kalantidi}. Popular aggregation methods such as Fisher Vectors, VLAD and RVD are adapted to encode deep features in the work of Ong et al. \cite{OngHB17}, Arandjelovic et al. \cite{NetVLAD} and Husain et al. \cite{HusainPAMI}. Tolias et al. \cite{Tolias2015} develop a hybrid scheme called Regional Maximum Activations of Convolutions (RMAC), where last convolutional layer features are first max-pooled over regions using a fixed grid. The regional features are PCA+whitened, L2-normalised and sum-aggregated to form an image signature. A modified version of RMAC that combines multi-scale and multi-layer feature extraction is introduced by Seddati et al. \cite{Seddati_2017}. The method of Jimenez et al. \cite{Jimenez_2017_BMVC} uses VGG16 and class activation maps to compute a global descriptor. Xu et al. \cite{SBA} propose a semantic-based aggregation approach where probabilistic proposals corresponding to special semantic content in an image are used to aggregate deep features. A hybrid deep feature aggregation method that unifies sum and weighted pooling to compute an image representation is presented by Pang et al. \cite{Unifying}.

The methods that use convolutional features from a CNN, trained for ImageNet classification, perform sub-optimally for IIR due to the CNN being tuned to optimise intra-class generalisation. Current methods aim to solve this limitation by finetuning the trained networks for retrieval tasks.

\subsection{Finetuned deep CNN image retrieval}
Arandjelovic et al. \cite{NetVLAD} introduce NetVLAD, where the last convolutional layer of VGG16 is followed by a generalized VLAD layer and the whole architecture is trained using weakly supervised triplet loss. Gordo et al. \cite{Gordo2017} build on the core RMAC network and train the deep image retrieval (DIR) architecture on the Landmarks dataset with triplet loss. Radenovic et al. \cite{GEM} propose to aggregate deep features using a generalized mean pooling (GEM) layer. More precisely, the GEM layer is added to the last convolutional layer of ResNet101 and the resultant architecture is trained on the Landmarks dataset to minimise contrastive loss. Two variants of GEM pooling are weighted generalized mean pooling (WGEM) \cite{WGEM} and attention-aware generalized mean pooling (AGEM) \cite{AGEM}. The GEM-based aggregation methods are related to our approach but are restricted to a specific Lp-norm family of functions. Our method provides a richer family of aggregation functions that are not possible with GEM-based approaches. Further discussion can be found in Section \ref{sec:gem_relation}.

Xu et al \cite{ASDA} develop an adversarial soft-detection-based aggregation (ASDA) network which pools deep features using adversarial detector and soft region proposal layer. Teichmann et al. \cite{D2R} propose to aggregate and match deep local features (DELF) based on regional aggregated selective match kernels (R-ASMK). Husain et al. \cite{REMAP} develop a deep architecture, Region-Entropy based Multi-layer Abstraction Pooling (REMAP), where multi-layer regional features are aggregated based on entropy-guided pooling. Recently, Revaud et al. \cite{GEMAP} propose to train GEM and RMAC networks using Listwise loss to improve retrieval accuracy.

In summary, while very significant progress has recently been achieved, the retrieval performance of state-of-the-art methods is still underwhelming, especially on challenging large-scale datasets where many distractor images resembling the query image, but depicting a different object, exist. We attribute this weakness to three main causes, which our present work address, namely: (1) CNN architectures, and in particular the design of the aggregation stage which is responsible for robust global descriptor generation, (2) training protocols resulting in sub-optimal learning of extreme cases, and (3) mapping of the deep representations into compact signatures, which may lead to the reduction of its discriminatory power. 

\begin{figure*}
	\centering
	\includegraphics[width=1.0\textwidth]{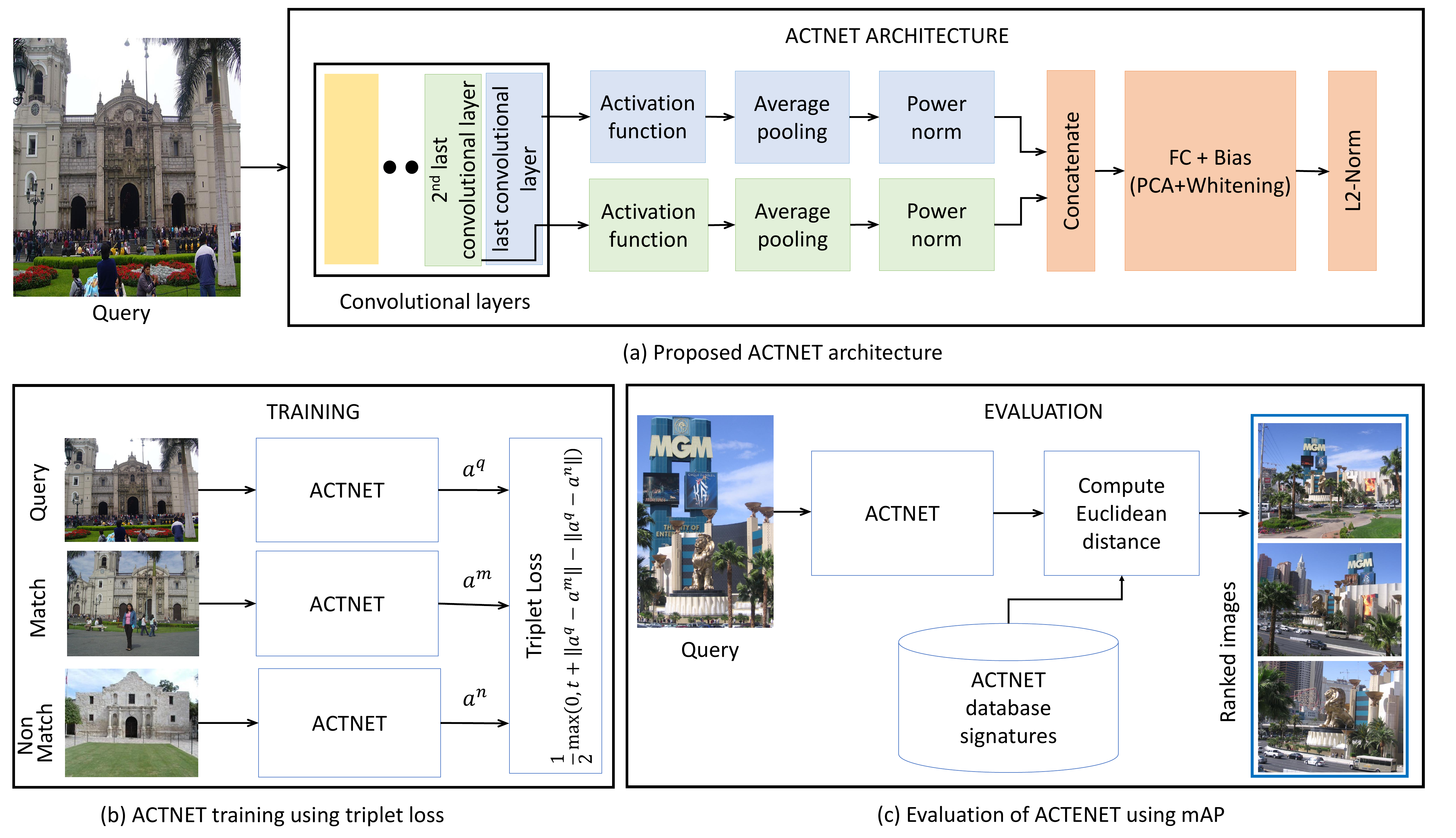}
	\caption{(a) Proposed ACTNET architecture with multi-stream activation based aggregation, (b) training of ACTNET using triplet loss and (c) evaluation of ACTNET on state-of-the-art datasets}
	\label{ARCH}
\end{figure*}

\section{ACTNET architecture} \label{sec:actnet}
In this section, we present the motivation, introduce details of our novel ACTNET architecture, describe its components and conduct an in-depth experimental evaluation of various elements of ACTNET. 

First, we present a novel learnable aggregation for deep descriptors. We observe that distinctive features (i.e. represented by strong CNN responses) are typically contaminated by the background noise, i.e. the signals (CNN responses) from weaker but numerous features. In the max-pooling aggregation, only the strongest response is retained in each channel, however there is no guarantee that the strongest response corresponds to the object of interest. In the average pooling, on the other hand, all responses are averaged within channels, leading to the contamination effect. Our aim is to maximise the signal-to-noise ratio (SNR) of the convolutional feature map before aggregation, i.e. we would like to amplify the signals originating from the actual object of interest in an image, relative to the background noise. In our design, this is achieved by passing filter responses from convolutional layer through a learnable, non-linear "amplifier" function, implemented as a parametric activation layer. We propose three differentiable amplification functions, and perform an in-depth experimental study to compare their behaviour. The results show that activation functions significantly enhance the robustness and discriminative power of deep features, leading to the state-of-the art retrieval performance.

Second, based on our parametric activation layer, we design a deep multi-stream architecture, called Activation Network (ACTNET), to optimise retrieval tasks. We note that various convolutional layers of a deep CNN act as powerful signal generators that extract dense local features at various levels of semantic abstraction and with varying extent of region (spatial) support. However, conventional aggregation methods, such as max- or average-pooling fail to combine such features effectively.  In ACTNET, multiple parametric activations deliver adaptive multi-stream aggregation, where a hierarchy of deep features from different convolutional layers are amplified using activation functions and optimally aggregated into a discriminative global signature. The ACTNET architecture is trained in an end-to-end manner, where the convolutional weights, activation parameters and PCA+Whitening weights are jointly trained with triplet loss. The key aspect is that ACTNET learns the activations across multiple layers during training to select and balance features reflecting complementary and distinct levels of visual abstractions, thus boosting retrieval performance. 

While current image retrieval systems exhibit good extraction speed on GPU-accelerated computers, our aim is to deliver an efficient and effective system in the hands of mobile users. To this end, we develop a low power, low latency and small memory footprint ACTNET suitable for running on mobile devices. The low-complexity Mobile-ACTNET achieves performance comparable to computationally expensive methods such as GEM \cite{GEM} or RMAC \cite{Gordo2017} with a five times faster descriptor extraction speed.

ACTNET, presented in Figure \ref{ARCH}a, consists of a baseline CNN followed by our aggregation network. Any CNN can serve as the base; we show the results on two distinct networks: the high-performance ResNext101 \cite{RESNEXT} and low-complexity MobileNetV2 \cite{MobileNet}. For these CNNs, only the convolutional layers are retained, with the fully connected layers for classification discarded. The outputs of the final convolutional blocks are each fed to separate ``aggregation streams'' that utilise our proposed activation layers. In each stream, the activation layer is followed by average pooling and power normalisation layers. The outputs of all streams are then combined with a concatenation layer. This is followed by a PCA+Whitening layer before the final L2-normalisation layer, producing a discriminative global descriptor for an image. The training procedure (shown in Figure \ref{ARCH}b) adopts a three stream siamese architecture where the image signatures extracted by each of the streams are jointly examined by the triplet loss function. The training procedure for the ACTNET is described in Section \ref{sec:actnet_train}. In the evaluation phase (Figure \ref{ARCH}c), an ACTNET signature is computed from a query image and compared against a database of pre-computed signatures using Euclidean distance. Based on the similarity scores, the database images are ranked and mean Average Precision (mAP) is computed.

\subsection{Learnable non-linear activation layer}
\label{sec:act_layer}
In this section, we present a novel CNN layer which we call the `activation layer' that provides the ability to learn and tune non-linear activation functions. The aim is to give greater emphasis to important responses from the object in an image and reduce the impact of background noise by non-linearly scaling values of the input tensors. Crucially, this non-linear scaling is tune-able during the learning process.

The fundamental difference between our work and state-of-the-art on developing adaptive activation functions is that we aim to improve the SNR of the final convolutional feature map before aggregation into a global descriptor while current work focuses on improving the speed and stability of the network. Our activation function non-linearly amplifies convolutional layer values, thereby improving the robustness and discriminative power of deep features. Recent works on using activation functions to help fast training of CNNs are \cite{Jarrett}, \cite{glorot}, \cite{goodfellow}, \cite{QIAN}, \cite{Agostinelli2014LearningAF}. The rectified linear activation (ReLU) function \cite{glorot} has made it easier to effectively train a CNN compared to activation functions such as sigmoid or tanh, by addressing the problem of vanishing gradients. Agostinelli et al. \cite{Agostinelli2014LearningAF} propose to use an adaptive linear piecewise model for the activation function. Farhadi et al. \cite{Farhadi2019ActivationAI} propose an adaptive ReLU function that provides a smoother scaling increase of activation values. Qian et al. \cite{QIAN} explore methods to linearly combine basic activation functions. 
%

We denote the non-linear activation function as $\phi(\mathbf{X};\theta_\phi,\mathbf{w}_\theta)$, where $\phi:\mathbb{R}^{W\times H\times D} \rightarrow \mathbb{R}^{W\times H \times D}$ is an operation that takes as input a tensor $\mathbf{X}$ of size $W\times H\times D$ and applies an associated real-valued function $\theta_\phi:\mathbb{R}\rightarrow\mathbb{R}$ to each element of the input tensor. Importantly, we consider parameterised functions $\theta_\phi$, where their respective parameters are denoted as a vector of real values: $\mathbf{w}_\theta$. More specifically, we have:
\begin{equation}
\phi(\mathbf{X};\theta_\phi,\mathbf{w}_\theta) = \left(\theta_\phi(x_{ijk}; \mathbf{w}_\theta) \right)^{W,H,D}_{i,j,k=1}
\label{eq:phi}
\end{equation}

\begin{figure*}
	\centering
	\includegraphics[width=1.0\textwidth]{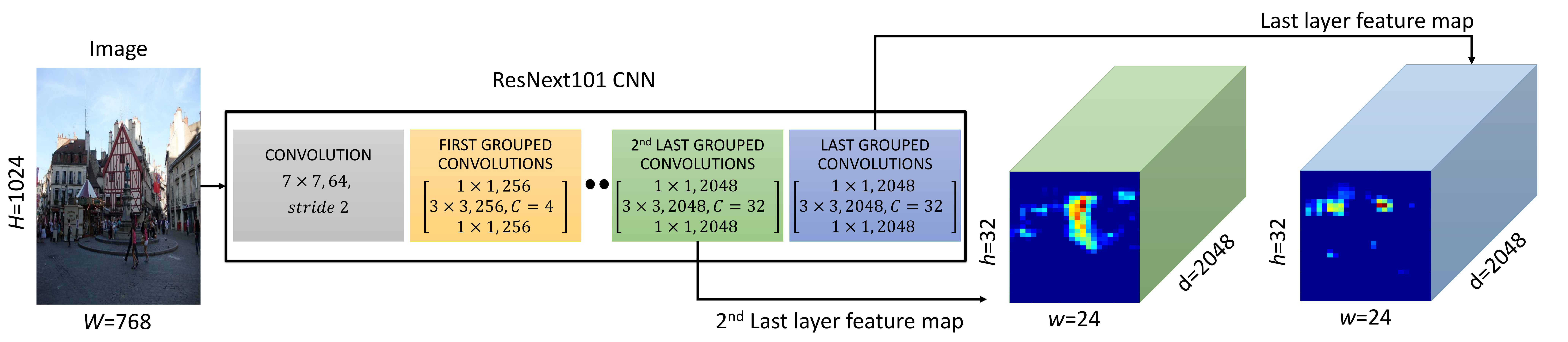} \\
	(a) An illustration of how the inputs to the activation functions are generated. The output tensors from the final convolutional layers are passed to the activation functions shown below.\\
	\includegraphics[width=1.0\textwidth]{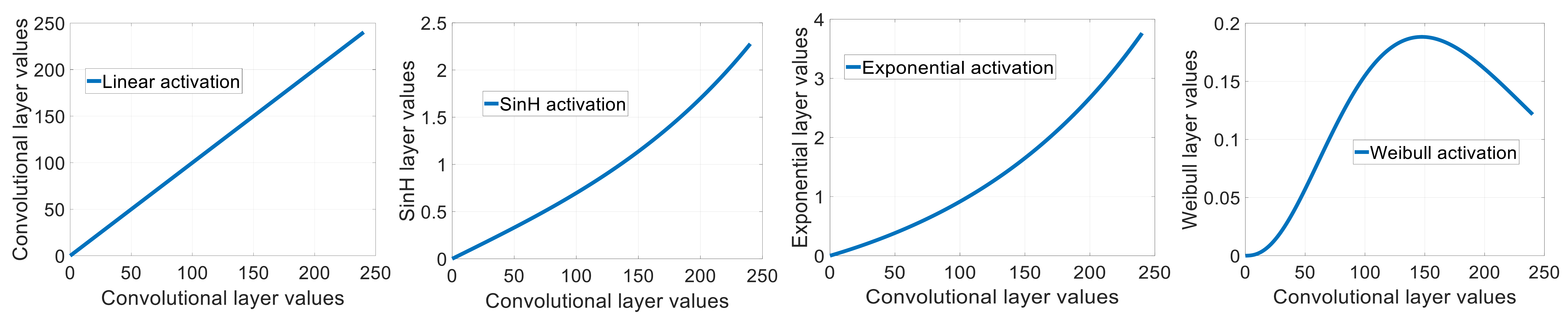} \\
	(b) The profile curves of the activation functions considered in this paper: Sine-Hyperbolic, Exponential and Weibull.
    \includegraphics[width=1.0\textwidth]{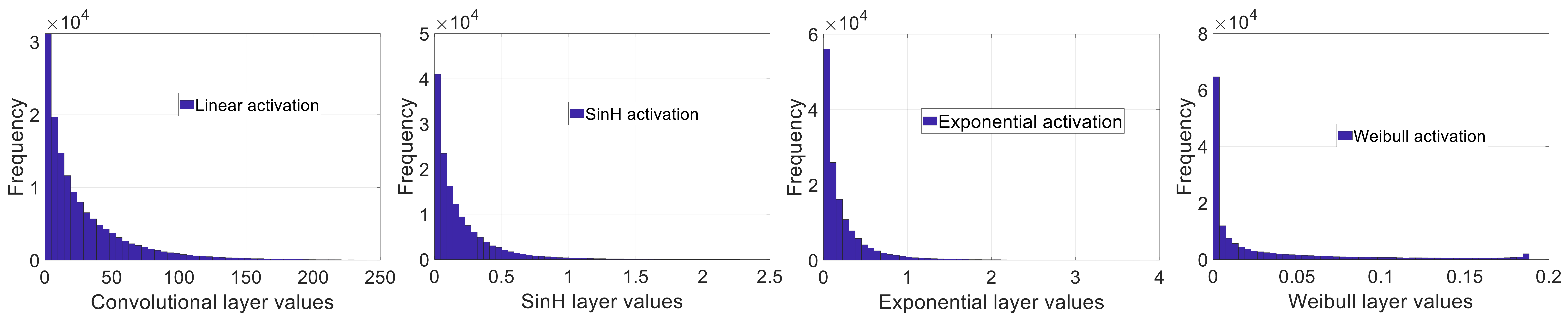} \\
	(c) Histograms of feature map values before and after passing through activation functions.
	\caption{Process of transforming convolutional layer values using the activation functions}
	\label{Activations-fig}
\end{figure*}

In this work, we have considered three different activation functions for $\theta_\phi$: Sine-Hyperbolic, Exponential and modified Weibull. These three are selected thanks to the following properties that they have in common:
\begin{itemize}
\item
Each activation function non-linearly amplifies the input values within a real-valued interval. This allows us to place greater emphasis on larger values (important activations), compared to a linear scaling. 
\item The non-linearity of the amplification is adjustable for learning purposes. As such, all the activation functions considered are parameterised. 
\item Finally, for use in a stochastic gradient descent framework, we require all the functions to be differentiable with respect to all their parameters. 
\end{itemize}
We next give the details of the three different activation functions used in ACTNET. For clarity and convenience, we will omit writing the vector of parameters, $\mathbf{w}_\theta$, when describing the function, writing it as $\theta_\phi(x)$ instead of $\theta_\phi(x;\mathbf{w}_\theta)$.

\subsection*{Sine-Hyperbolic function (SinH)}
The first activation function we consider is the monotonically increasing Sine-Hyperbolic.
\begin{equation}
    \theta_\phi(x) = \alpha\sinh(\beta x)
\end{equation}
where $\alpha$ and $\beta$ are the scaling parameters. 
The $\beta$ parameter determines the strength of the non-linear re-weighting of the original activation values. As such, it effectively controls the influence that higher valued features have on the average pooling operation performed later on. The $\alpha$ parameter has two important roles: (1) normalising the values of the scaled activations, (2) controlling the information flow from each stream resulting in a discriminative global descriptor.
The set of parameters associated with this activation function is: $\mathbf{w}_{\theta} = (\alpha, \beta)$.

Since the Sine-Hyperbolic function is differentiable, the chain-rule can be readily applied to obtain the partial derivatives required for back-propagation:

\begin{eqnarray}
\frac{\partial \theta_\phi}{\partial x} & = & \alpha \beta \cosh(\beta x ) \\
\frac{\partial \theta_\phi}{\partial \alpha} &= & \sinh(\beta x) \\
\frac{\partial \theta_\phi}{\partial \beta} & = & \alpha x \cosh( \beta x )
\end{eqnarray} 

\subsection*{Exponential function (Exp)}
The second activation function proposed to improve the SNR of feature maps is based on the exponential function: 
\begin{equation}
    \theta_\phi(x) = \alpha(\exp(\beta x)-1)
\end{equation}
where $\alpha$ and $\beta$ are the scaling parameters, which play analogous roles as the parameters in the SinH activation function. Thus, the set of parameters associated with this function is given as:
 $\mathbf{w}_{\theta} = (\alpha, \beta)$.
The partial derivatives used for learning for the exponential function are as follows:
\begin{eqnarray}
\frac{\partial \theta_\phi}{\partial x} & = & \alpha \beta \exp(\beta x ) \\
\frac{\partial \theta_\phi}{\partial \alpha} & = & \exp(\beta x )-1 \\
\frac{\partial \theta_\phi}{\partial \beta} & = & \alpha x\exp( \beta x )
\end{eqnarray}

\begin{figure*}
	\centering
	\includegraphics[width=0.9\textwidth]{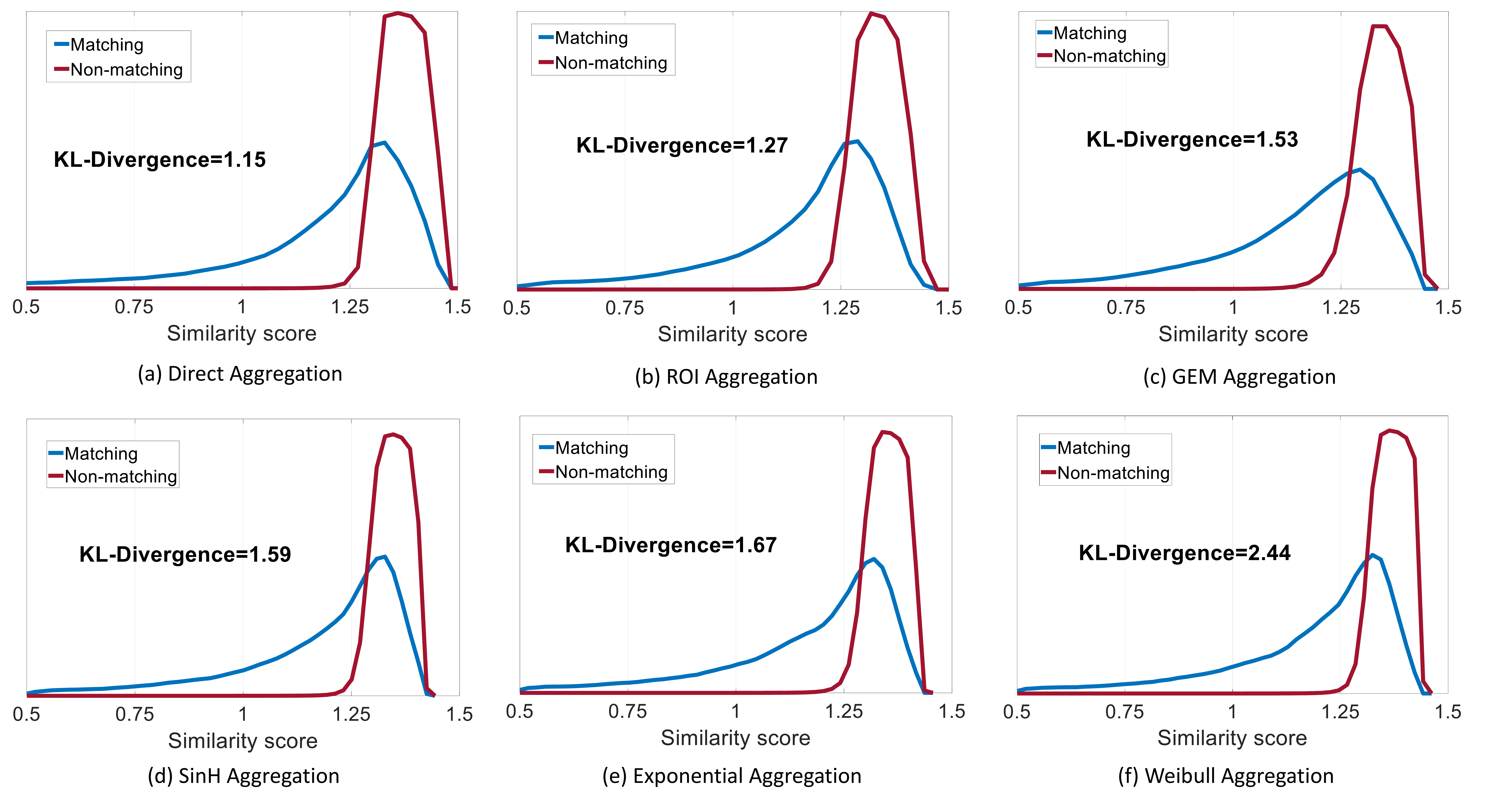}
	\caption{Histograms of Euclidean distances between matching and non-matching descriptors, for different aggregation methods. These are for single aggregation-stream networks (c.f. Section \ref{sec:actnet_arch}).}
	\label{fig:KLDIV}
\end{figure*}

\subsection*{Modified Weibull function (WB)}
One issue with the monotonically increasing functions such as SinH and Exp is that these functions are blinded by the strong activations of convolutional feature map. To solve this issue, we propose to transform the input values by applying a modified Weibull function:
\begin{equation}
\theta_\phi(x) = \left(\frac{ x}{\alpha}\right)^{\beta-1}\exp(-(x/\gamma)^\zeta)
\end{equation}
with parameters: $\mathbf{w}_{\theta} = (\beta, \alpha,\gamma,\zeta)$. Here $\beta$ is known as the shape parameter and determines where the peak of the activation function will be. The Weibull function is a product of two terms, one term increasing at polynomial rate $( x/\alpha)^{\beta-1}$, another term decreasing at an exponential rate $\exp(-(x/\gamma)^\zeta)$. For small enough values, the polynomial term dominates, thus the effect of this activation function is to non-linearly increase these values. However, eventually the inverse exponential term starts to dominate, thereby reducing the output value as the input value further increases. 
The point $x_0$ where activation values change from being increased to being decreased can be found by solving $\partial \theta_\phi/\partial x = 0$ for $x$:
\begin{equation}
x_0 = \gamma \left( \frac{\beta - 1}{\zeta} \right)^{1/\zeta}
\end{equation}
When we consider how the Weibull function changes the result of the average pooling operation later, we find that two fundamentally different forms of re-weighting occur: 1) tensor values before $x_0$ will have a non-linearly greater influence (polynomial rate), the closer they are to $x_0$; 2) tensor values after $x_0$ are effectively ``reversed'', where they exert an exponentially decreasing influence as they get larger.

The partial derivatives used for back propagation are:
\begin{eqnarray}
\frac{\partial \theta_\phi }{\partial x} & = & \left(\frac{\beta-1}{\alpha}\right)
\left(\frac{ x}{\alpha}\right)^{\beta-2}\exp(-(x/\gamma)^\zeta) - \\
& & \frac{\zeta}{\gamma^\zeta}\left(\frac{x}{\alpha}\right)^{\beta-1} x^{\zeta - 1}\exp(-(x/\gamma)^\zeta)\\
\frac{\partial \theta_\phi }{\partial \beta} & = & \log\left(\frac{x}{\alpha}\right)\left(\frac{ x}{\alpha}\right)^{\beta-1}\exp(-(x/\gamma)^\zeta) \\
\frac{\partial \theta_\phi}{\partial \alpha} & = & (1-\beta)\left(\frac{x^{\beta-1}}{\alpha^\beta}\right)\exp(-(x/\gamma)^\zeta) \\
\frac{\partial \theta_\phi}{\partial \gamma} & =& \zeta\left(\frac{x^{\beta+\zeta-1}}{\alpha^{\beta-1}\gamma^{\zeta+1}}\right)\exp(-(x/\gamma)^\zeta)\\
\frac{\partial \theta_\phi}{\partial \zeta} & =&  \left(\frac{x}{\alpha}\right)^{\beta-1}\left(-\frac{x}{\gamma}\right)^\zeta \log\left(\frac{x}{\zeta}\right)\exp(-(x/\gamma)^\zeta)
\end{eqnarray}

The above three activation functions are integrated into the ACTNET architecture in Section \ref{sec:actnet_arch}. However, we will first analyse how different activation functions affect the image features in the next section.

\begin{figure*}
	\centering
	\includegraphics[width=0.8\textwidth]{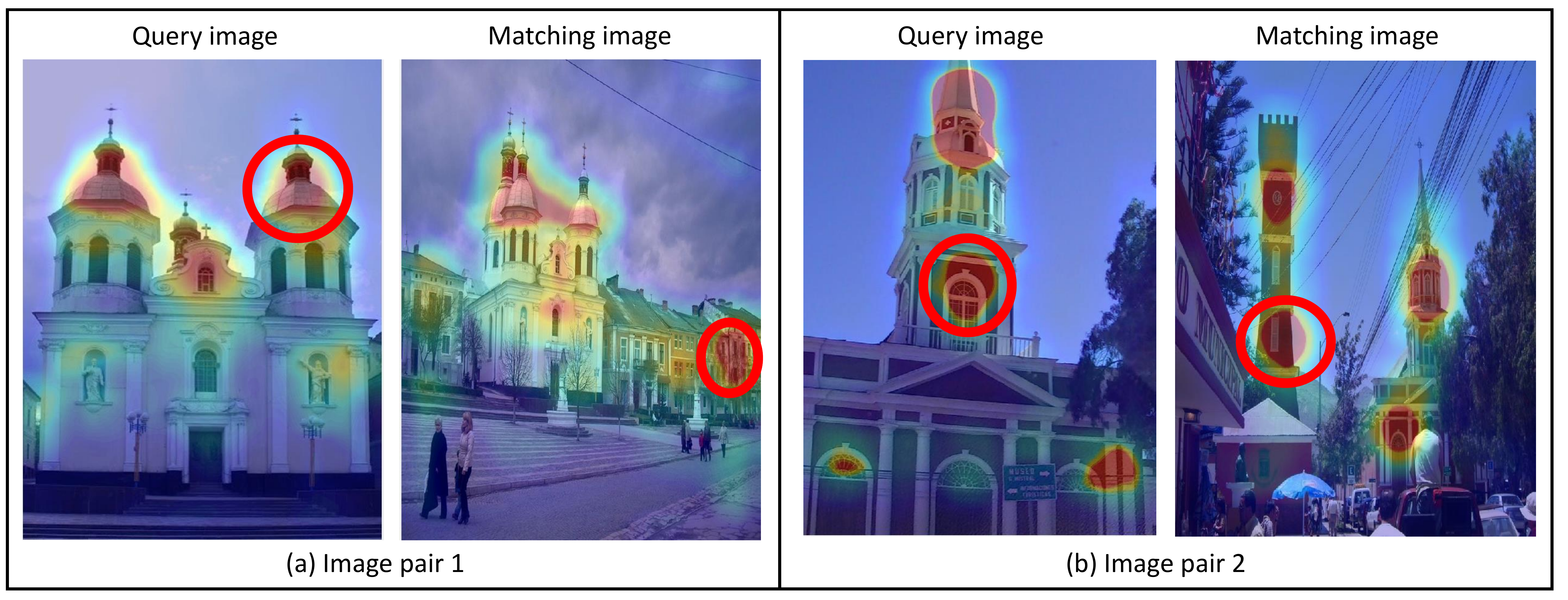}
	\caption{Visualisation of feature maps of the last convolutional layer for two pairs of images: each comprising of a query image and a matching image. The strongest responses are indicated by red circles.}
	\label{fig:WBact}
\end{figure*}

\subsection{Analysis of activation functions}
\label{sec:activ_func_choices}

Figure \ref{Activations-fig} demonstrates the process of applying activation functions to the features. Firstly, an input tensor for the activation function is obtained by extracting selected convolutional layer output from a base CNN (here ResNext101). The elements of this tensor fall in some non-negative range (Figure \ref{Activations-fig}a). We now apply different functions for non-linearly scaling the values of tensor. Transformations of input values by the activation functions considered are shown in Figure \ref{Activations-fig}b. Here, it can be seen that the three activation functions have different shapes, providing different degrees of non-linear amplification depending on the signal level and the associated parameters. Importantly, the addition of the activation layer before aggregation gives us a significant improvement over direct aggregation where the deep features are linearly scaled and aggregated (refer Section \ref{sec:experminets}). The SinH function provides a non-linear increase that is in between the linear curve and exponential curve. The Exponential function scales up the input values more steeply than SinH. Finally, we see that the Weibull function provides us with a robust amplification, increasing values up to a certain point, before its activation values starts decreasing; the rationale here is that we aim to equalise the power of few strong activations corresponding to prominent and distinctive image features. This “equalisation” is crucial as we cannot know - at the extraction and aggregation stage - which features correspond to “foreground” and which to “background”. This depends on the query and the other image being matched, and for some queries the strong activations may correspond to the background clutter, hence we do not want any activation to significantly “dominate” the aggregation process. In summary the Weibull function gives us two benefits - attenuation of the weaker activations and equalisation of the stronger ones.  

The distributions of tensor values before and after passing through the activation functions for a sample image are presented in Figure \ref{Activations-fig}c. It can be observed that activation functions significantly improve the SNR of feature maps by squeezing the majority of noisy responses to insignificantly small values. Importantly, the Weibull function results in a distribution with a large peak within the first bin showing the noise reduction phenomenon as well as a small peak in the final bin of the distribution showing the ability to equalise strong activations.

The application of activation functions to the feature map values modifies the underlying distributions of similarity scores for matching and non-matching pairs of images. To demonstrate the benefits of the proposed activation layer, we compute the class-separability between histograms of distances for matching and non-matching image pairs.
The descriptors are extracted using the following aggregation approaches:
\begin{enumerate}
    \item Direct aggregation (DA) \cite{BabenkoSCL14}: the features from the last convolution layers are aggregated using average pooling.
    \item Region of Interest based aggregation (ROIA) \cite{Gordo2017}: the features are first max-pooled across several multi-scale overlapping regions. The regional descriptors are aggregated using sum pooling. 
    \item GEM aggregation (GEMA) \cite{GEM}: the convolutional features are aggregated using generalized mean pooling.
    \item SinH aggregation (SinHA): the features are transformed using the Sine-Hyperbolic activation layer before aggregation using average pooling.
    \item Exponential aggregation (ExpA): the features are passed through the Exponential activation layer before average pooling.
    \item Weibull aggregation (WBA): the features are transformed using the Weibull activation layer before average pooling.
\end{enumerate}
In all the aforementioned methods, the aggregated signatures are then PCA+Whitened and L2-normalised to form a global descriptor. We then compute $P(s/m)$ and $P(s/n)$ as the probability of observing a Euclidean distance $s$ for a matching and a non-matching descriptor pair $m$ and $n$ respectively. The separability between $P(s/m)$ and $P(s/n)$ across values of $s$ is computed using KL-Divergence (KLD). It can be observed from Figure \ref{fig:KLDIV} that the aggregation methods based on activation functions, SinHA (KLD=1.59), ExpA (KLD=1.67), WBA (KLD=2.44), provide much better separability between matching and non-matching distributions when compared to DA (KLD=1.15) and ROIA (KLD=1.27). There is also an improvement compared to the parametric GEM aggregation (KLD=1.53). It can also be observed that the Weibull aggregation achieves the highest KL-Divergence of 2.44. Importantly, we find that an increase in KL-divergence correlates strongly with the increase in retrieval accuracy, as shown in Section \ref{sec:experminets}.

In Figure \ref{fig:WBact}, we visualise why it is important to apply a robust activation function before aggregation, in our case the modified Weibull function. We show the last convolutional layer responses for two pairs of images: each comprising of a query image (left) and a matching image (right). The strongest responses are indicated by red circles. The max-pooling method of \cite{Razavian} generates global descriptors from a query image and matching image that have a very low similarity score (Euclidean distance 1.42); this is because the maximum activation for the matching image (right) is from the background and not the actual object to be retrieved. The average-pooling \cite{BabenkoSCL14} method alleviates the problem of max-pooling, however the matching descriptor is now contaminated with significant amount of background responses resulting in a low similarity score (Euclidean distance 1.39). The application of SinH or Exponential functions to the input tensor significantly improves the signal-to-noise ratio thereby increasing the similarity score between global descriptors (Euclidean distance 1.27 and 1.21). The additional benefit of the Weibull activation function is that it balances the influence of the strong activations, thus providing the best similarity score (Euclidean distance 1.08). 

\subsection{Image retrieval ACTNET architecture}
\label{sec:actnet_arch}
We will now formally describe the ACTNET architecture. 
First, let $\mathbf{x} \in \mathbb{R}^{W\times H \times 3}$ denote an RGB input image of resolution $W \times H$.
Next, for the selected base CNN, given an input image $\mathbf{x}$, suppose we are interested in using $K$ of its convolutional layer outputs. Let us denote these $K$ convolutional layers as functions: $f_1(\mathbf{x}), f_2(\mathbf{x}),..., f_K(\mathbf{x})$.
Associated with each of these functions are their own set of learnable parameters, which are optimised via back-propagation using the partial derivatives $\partial \theta_\phi/\partial x$ given in Section \ref{sec:act_layer}. These convolutional outputs are then be passed as inputs to their own ``aggregation streams'' as described next.

\subsection*{Aggregation stream}
The aim of the aggregation stream is to perform non-linear scaling (activation), followed by aggregation of convolutional features and finally deactivation scaling. First, the input tensor $\mathbf{X} \in \mathbb{R}^{W\times H \times D}$ is transformed using the activation function $\phi$ (Eq. \ref{eq:phi}) resulting in the output tensor $\mathbf{Y} \in \mathbb{R}^{W\times H \times D}$
\begin{equation}
    \mathbf{Y}=\phi(\mathbf{X}; \theta_\phi, \mathbf{w}_\theta)
\end{equation}
We next apply global average pooling denoted by $P(\mathbf{Y})$:
\begin{equation}
    P(\mathbf{Y}) = \left(\frac{1}{WH}\sum^W_i\sum^H_j y_{ijk} \right)_{k = 1}^D
\end{equation}
The output of the global average pooling operation is the vector $\mathbf{z} = P(\mathbf{Y})$, which we normalise to balance the severity of increase in values due to the non-linear scaling process of the activation layer. In our work, we have found that the choice of fractional power achieves best results. The power normalisation function is denoted as $\psi:\mathbb{R}^D\rightarrow\mathbb{R}^D$, with the rule:
\begin{equation}
    \psi(\mathbf{z}) = \lambda z_1^{p},\lambda z_2^{p} ,..., \lambda z_D^{p}
\end{equation}
where $\lambda, p$ are learnable scaling parameters.

To bring it all together, we can now define the full aggregation stream as the function: $\Phi:\mathbb{R}^{W \times H\times D} \rightarrow \mathbb{R}^D$, with the rule:
\begin{equation}
\Phi(\mathbf{X}) = \psi( P(\phi(\mathbf{X}; \theta_\phi, \mathbf{w}_{\theta})) )
\label{eq:phi_general}
\end{equation}

To aid the following discussion when multiple streams are considered, we gather all the learnable parameters described above into a set denoted as: 
\begin{equation}
\mathbf{\eta} = \{\mathbf{w}_\theta, p, \lambda\}
\label{eq:stream_param}
\end{equation}

\subsection*{Global descriptor generation}
We now describe how the global descriptor of an image $\mathbf{x}$ will be generated using convolutional features from the base CNN. 

As noted previously, we are interested in using $K$ convolutional layer outputs from the base CNN. Each such convolutional layer gives rise to a separate aggregation stream of form Eq. \ref{eq:phi_general}. We denote each of the $K$ separate streams as: $\Phi_1(f_1(\mathbf{x})),\Phi_2(f_2(\mathbf{x})),..., \Phi_K(f_K(\mathbf{x}))$. Their respective learnable parameters (Eq. \ref{eq:stream_param}) are denoted as: $\mathbf{\eta}_1,\mathbf{\eta}_2,...,\mathbf{\eta}_K$. The global descriptor is then constructed by concatenation of the outputs of all aggregation streams:
\[
\mathbf{b} = [\Phi_1( f_1(\mathbf{x})),\Phi_2( f_2(\mathbf{x})),..., \Phi_K( f_K(\mathbf{x})) ]
\]
The concatenated feature vector $\mathbf{b}$ is then passed to a PCA and whitening layer followed by  a L2-normalisation layer, which produces the final global descriptor used for image retrieval.

\subsection*{Relation to state-of-the-art aggregation approaches}
\label{sec:gem_relation}
In this section we conduct an in depth study to understand the differences between three aggregation methods: ACTNET, GEM and REMAP.

In GEM \cite{GEM}, the features from the last convolutional layer are aggregated using generalized mean pooling. We can disassemble the aggregation method of GEM into a sequence of activation-aggregation-deactivation operations as well. GEM can be considered as a specific scaled Lp-norm, where the activation and deactivation are polynomial (degree say $p$) and its inverse respectively. In fact, for the past 20 years, Lp-norm has been used as a method for aggregation, and to our knowledge, simply because it improved results. In our work, we break the link between the activation and deactivation stages and propose a novel class of functions that are fundamentally different and more flexible, enabling transformations of activations in ways not possible by any Lp-norm. An example is our modified Weibull function specifically designed to guard against being blinded by strong activations due to its non-monotonic nature within the activation stage, which leads to dramatic improvements in performance accuracy. The new insight is that the activation stage transfer functions can and should be designed to optimise signal-to-noise ratio for activations, and there is no benefit for the activation and deactivation to be ‘coupled’ as in GEM.

In REMAP \cite{REMAP}, the features from multiple CNN streams are aggregated using entropy-guided ROI pooling. In the REMAP design, aggregation streams are controlled by KL-divergence during training. Essentially, REMAP learns offline which fixed regions from which layers are most informative and has no ability to adjust to particular activations in individual images. While ACTNET also aggregates features from multiple layers, it uses a diametrically different (and novel) approach to learn and control information flow via the network, by shaping individual activations to optimise SNR and the information retained in the aggregated descriptor. This means that we can forgo the arbitrarily pre-defined regions of REMAP, and also results in very substantial performance gains over REMAP as shown in Section \ref{sec:GEM,REMAP,ACTNET}

\subsection{End-to-end training of ACTNET}
\label{sec:actnet_train}
A vital aspect of the ACTNET is that all its components are considered to represent differentiable operations. Our novel activation layer is differentiable with parameters that can be optimised during training. The average pooling layer and the normalisation layer are also differentiable. The PCA+Whitening projection is implemented as a Fully Connected layer (for the projection with whitened eigenvectors) with bias (for mean subtraction), with weights that can be freely optimised. In conclusion, ACTNET is an end-to-end architecture which can learn the convolutional filter weights, activation parameters and PCA+Whitening parameters, using Stochastic Gradient Descent (SGD) on triplet loss function during training. 

We will now show how to build and train ACTNET using ResNext101 \cite{RESNEXT} as the base CNN. The first phase of training starts by fine-tuning ResNext101 (trained on ImageNet) on the GLRD dataset (refer Section \ref{sec:experminets}) using classification loss. In the second phase, we remove the last three layers of the tuned ResNext101 and add the activation layers, average pooling layers and power normalisation layers to the last two convolutional blocks. The outputs from the normalisation blocks are concatenated and a PCA+Whitening layer is added to form the final ACTNET. We then adopt a three stream siamese architecture to train the warmed-up ACTNET using triplet loss. For siamese network training, a dataset of $R$ triplets, each comprising of a query image, a matching image and a non-matching image is considered. More precisely, let $a^q$ be an ACTNET representation of a query image, $a^m$ be a representation of a matching image and $a^n$ be a representation of a non-matching image. The triplet loss can be defined as: 
\begin{equation}
L= \frac{1}{2} \; \max\left(0,t + || a^q - a^m ||^2 -|| a^q - a^n ||^2\right),
\label{eq:loss}
\end{equation}
where $t$ controls the margin. During training, the aim of the triplet loss is to adjust the ACTNET parameters such that the distance between $a^q$ and $a^m$ reduces and distance between $a^q$ and $a^n$ increases.

Given an query image at test time, ACTNET produces a robust and discriminative $D=4096$ dimensional image representation, well-suited for image retrieval.

\subsection{Low complexity ACTNET}
The ACTNET systems based on high-performance CNNs such as VGG, ResNet101, ResNext101 and DenseNet require significant amount of computational resources beyond the capabilities of mobile devices. In this section, we develop a low complexity ACTNET to effectively maximise retrieval performance while being mindful of limited resource scenarios. The overall architecture consists of a  MobileNetV2 base CNN followed by our aggregation system. The procedure to create the low complexity ACTNET (Mobile-ACTNET) is as follows: we remove the last pooling layer, prediction layer and loss layer of MobileNetV2 (trained on ImageNet). The outputs of the final two convolutional layers are then passed to activation layers, average pooling layers, power normalisation layers and a concatenation layer, generating a global descriptor. Finally, a siamese network is adopted to train Mobile-ACTNET on the GLRD dataset using triplet loss.

Given a query image at test time, Mobile-ACTNET generates a 2560-dimensional image signature. The dimensionality of the final signature is reduced using PCA+Whitening transformation, thereby reducing memory requirements and increasing the matching speed. 

In all the following experiments, we refer to ResNext-based ACTNET as "ACTNET" and MobileNetV2-based ACTNET as "Mobile-ACTNET".

\begin{figure*}
	\centering
	\includegraphics[width=\textwidth]{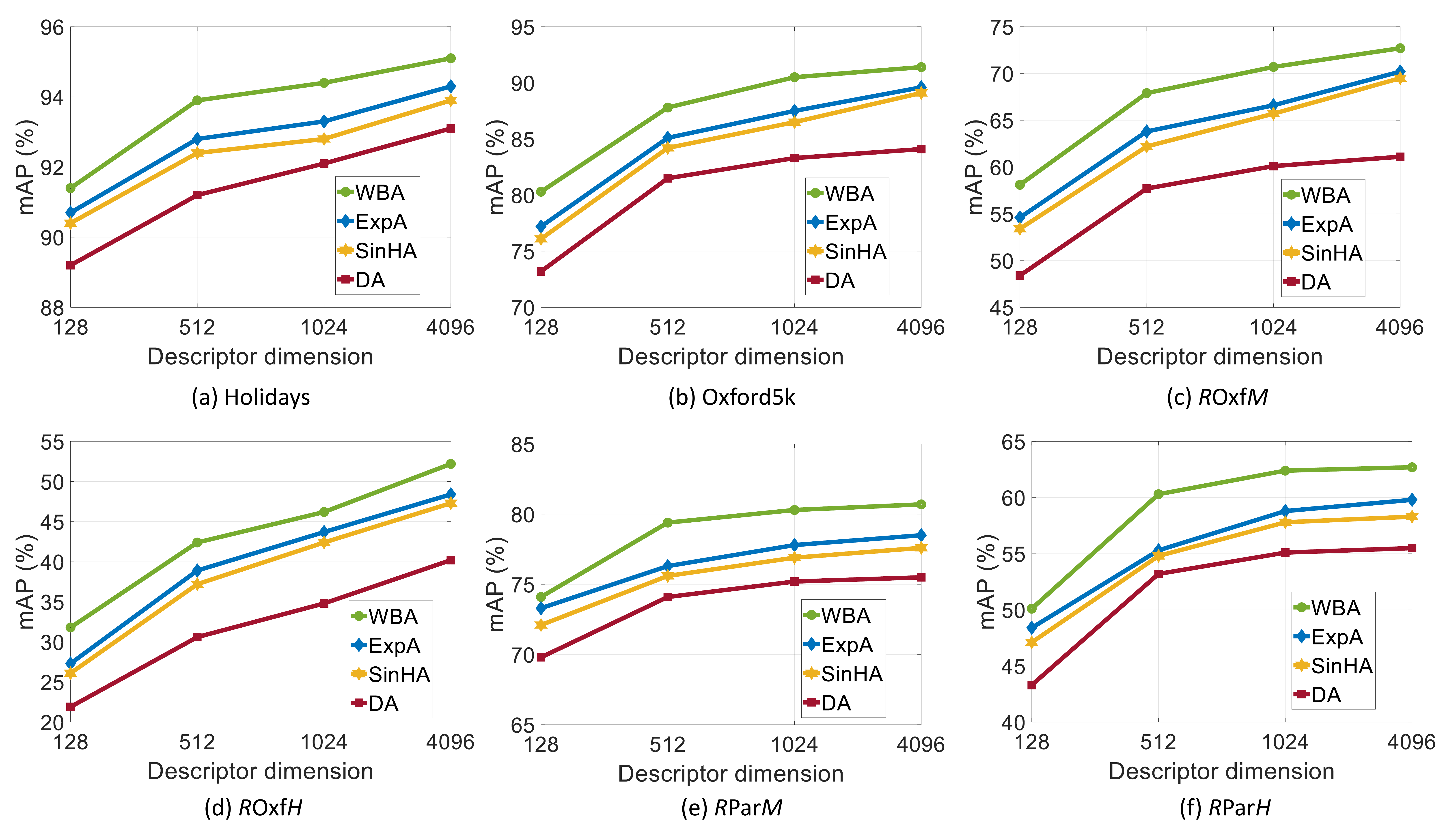}
	\caption{Comparison of retrieval performance with different aggregation methods: Direct aggregation (DA),  SinH aggregation (SinHA), Exponential aggregation (ExpA) and Weibull aggregation (WBA) on
		(a) Holidays, (b) Oxford5k, (c) $R$Oxf$M$, (d) $R$Oxf$H$, (e)$R$Par$M$, (f)$R$Par$H$, (all results in mAP(\%));}
	\label{RESULTS_ACT}
\end{figure*}

\section{Experimental results} \label{sec:experminets}
In this section we detail the implementation details of our architecture and give qualitative and quantitative results to validate our method. We first describe the experimental setup which includes the training parameters, training and test datasets and evaluation protocols. We then discuss the importance of the novel components of ACTNET, namely the robust activation function and multi-stream aggregation. Finally, we compare the proposed method to the state-of-the-art, showing significant improvements in retrieval performance across the board. 

\subsection{Training setup}
We train ACTNET on a subset of Google Landmarks dataset (GLD) \cite{Delf}, which contains 1 million images depicting 15K unique landmarks (classes). In GLD the number of images per class is highly unbalanced, some classes contain thousands of images, while for around 8K classes only 20 or fewer images are present. Furthermore, the GLD contains a non-negligible fraction of images unrelated to the landmarks. For this reason, we preprocess the GLD using the RVDW global descriptor and SIFT based RANSAC \cite{REMAP}, to obtain a clean dataset containing 200K images belonging to 2K classes. Finally, we remove all images that overlap with the test datasets. We refer this clean dataset as Google Landmarks Retrieval dataset (GLRD). We randomly select 1500 classes (150k images) for training and 500 classes (50k images) for validation.

We adopt a siamese architecture and train ACTNET on the GLRD dataset using triplet loss. The objective function is optimised by Stochastic Gradient Descent (SGD) with learning rate 1e-03, weight decay of 5e-04, momentum 0.9 and triplet loss margin 0.1. Each triplet contains a query, a matching and a hardest non-matching image and the size of each image is fixed to 1024$\times$768. To ensure that the sampled triplets incur loss and help in the training process, we extract the global descriptors from GLRD using the current model, and sample 5K triplets. All the triplets that cause a loss (the distance between query descriptor and non-matching descriptor is within margin 0.1 of the distance between query descriptor and matching descriptor) are sent to the network for training. After each epoch (training on 5K triplets), our algorithm generates a new set of 5K triplets using the current model for the following epoch. The training process is performed for at most 20 epochs. We overcome the TITAN X GPU memory limitation of 11 GB by processing one triplet at a time and updating the gradients after every 64 triplets. 

The initial values for SinH and Exp activation layers are: $\alpha=3$, $\beta=0.01$. The initial values for modified Weibull layer are: $\alpha=100$, $\beta=3.5$, $\gamma=80$ and $\zeta=1.5$. 

\begin{table*}[t]
\caption{Importance of multi-stream aggregation of convolutional features}
\centering
\label{Multimap}
\begin{tabular}{|c|c|c|c|c|c|c|c|}
\hline
 & Holidays & Oxford5k & $R$Oxf$M$ & $R$Oxf$H$ & $R$Par$M$ & $R$Par$H$ & MPEG \\ \hline \hline
SSA & 94.1 & 88.9 & 68.9 & 45.5 & 78.9 & 59.8 & 78.9 \\ \hline
MSA & 95.1 & 91.4 & 72.7 & 52.2 & 80.7 & 62.7 & 82.1 \\ \hline
\end{tabular}
\end{table*}

\begin{figure*}
	\centering
	\includegraphics[width=\textwidth]{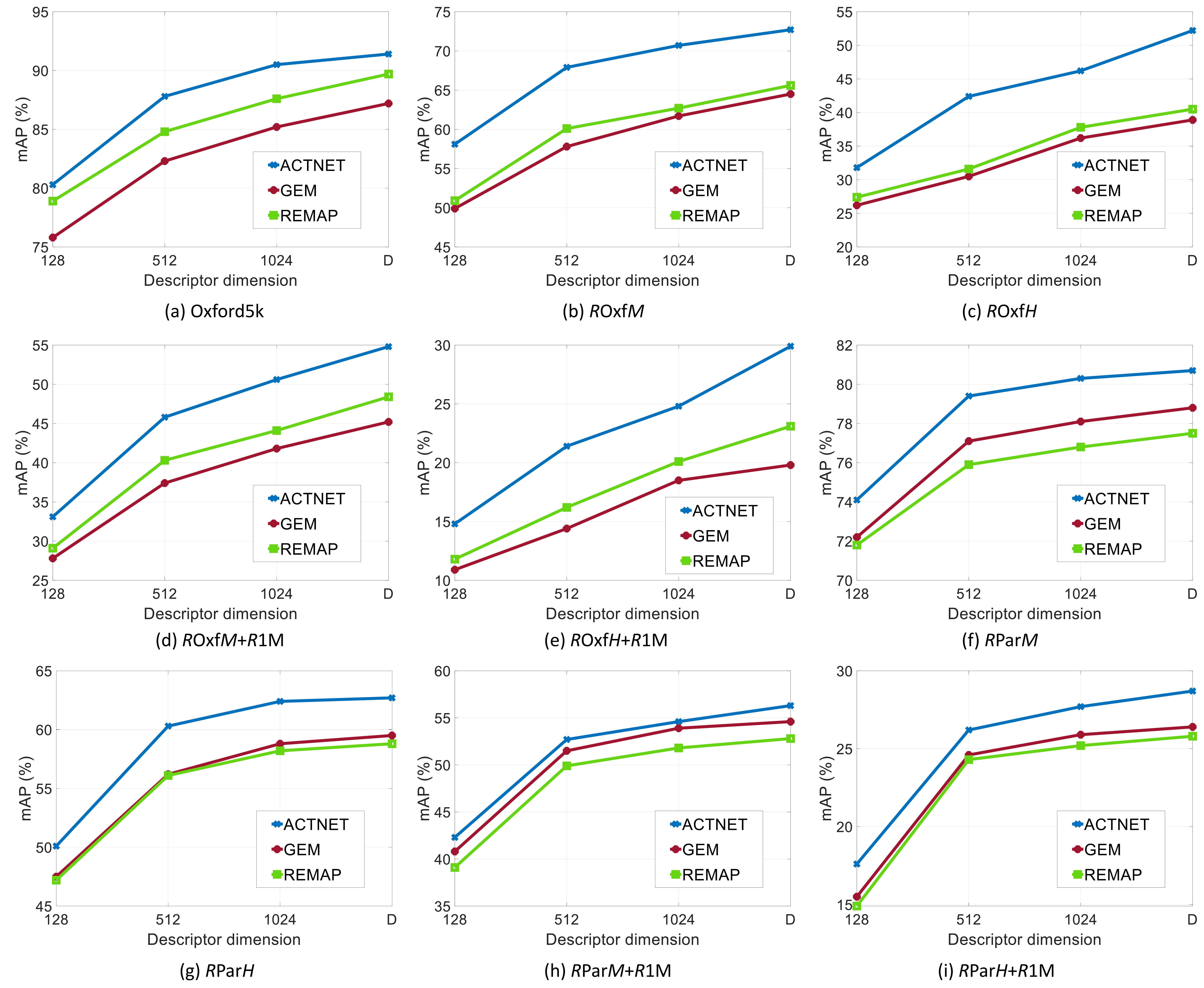}
	\caption{ACTNET comparison with GEM and REMAP networks
			(a) Oxford5k, (b) $R$Oxf$M$, (c) $R$Oxf$H$, (d) $R$Oxf$M$+$R$1M, (e) $R$Oxf$H$+$R$1M, (f) $R$Par$M$, (g) $R$Par$H$, (h) $R$Par$M$+$R$1M, (i) $R$Par$H$+$R$1M, (all results in mAP(\%));}
	\label{REMAP_vs_RMAC}
\end{figure*}

\subsection{Test datasets and evaluation protocol}
The original Oxford5k dataset contains 5063 high-resolution images, with a subset of 55 images used as queries. In \cite{ROXF}, Radenovic et al. revisited the original Oxford dataset to correct annotation errors, add 15 new challenging queries and introduce new evaluation protocols. The revisited Oxford dataset ($R$Oxf) comprises of 4993 images with 70 queries used for evaluation. 

The revisited Paris dataset ($R$Par) is a modified version of Parsi6K dataset: there are 70 query images with 6322 database images. 

The Holidays dataset \cite{JegouHE} consists of 1491 Holidays pictures, 500 of which are queries. Following standard procedure \cite{Gordo2017}, we manually correct the orientation of the images.

The Motion Picture Experts Group (MPEG) dataset is a collection of 35K images from five image categories: (1) Graphics, (2) Paintings, (3) Video frames, (4) Landmarks and (5) Common objects. A total of 8313 queries are used to measure retrieval performance.

To evaluate system performance in a more challenging scenario, the Holidays, Oxford5k, $R$Oxf and  $R$Par datasets are augmented with a distractor set containing 1 million non-matching images ($R$1M).
The retrieval performance for all datasets is computed using mean Average Precision (mAP). We follow the standard protocol for $R$Oxf and $R$Par datasets and report results using medium and hard setups referred as $R$Oxf$M$, $R$Par$M$, $R$Oxf$H$ and $R$Par$H$.

\begin{figure*}
	\centering
	\includegraphics[width=0.6\textwidth]{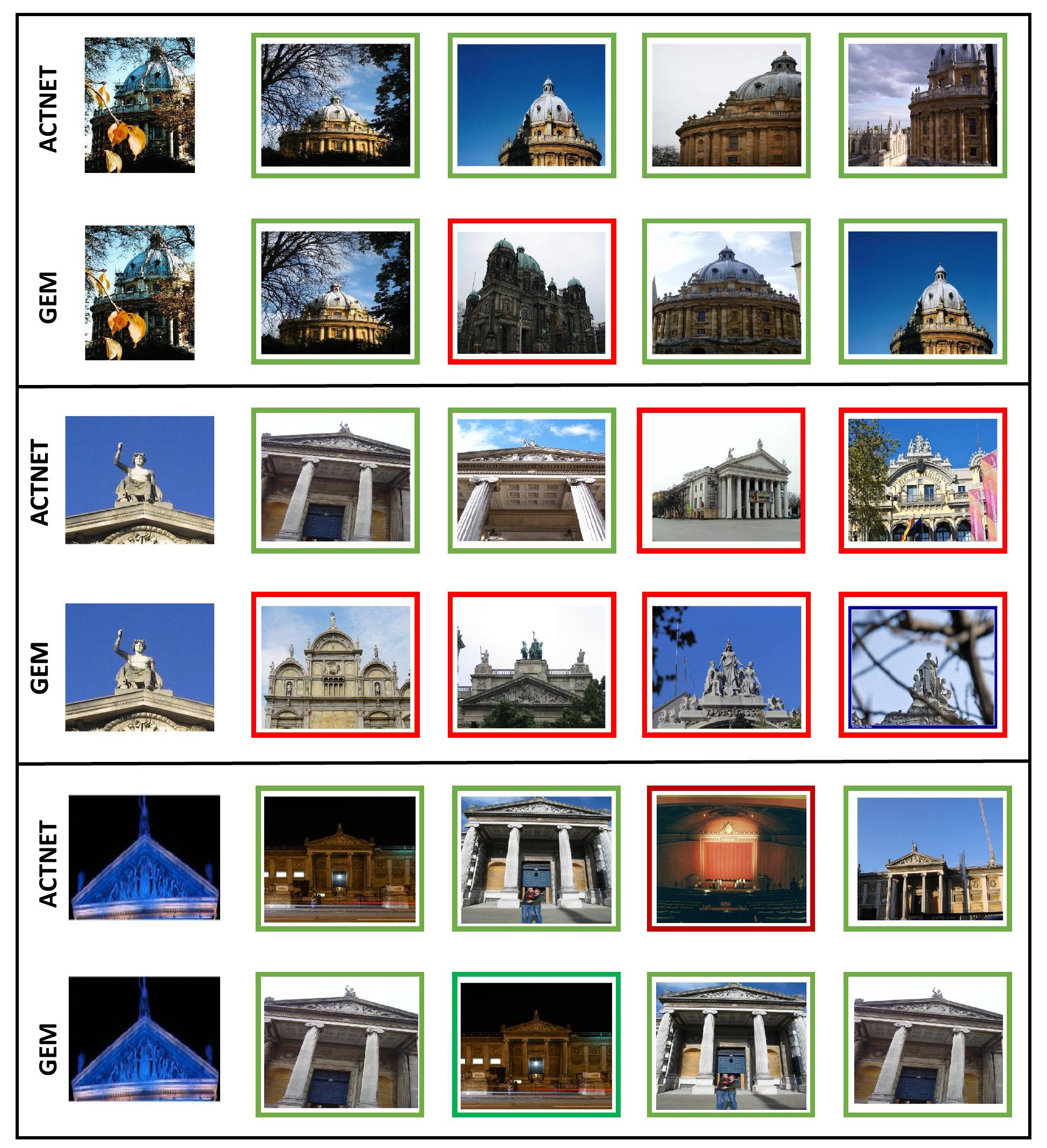}
	\caption{Top four retrieved images for ACTNET and GEM on $R$Oxf dataset. On the plot, the correctly retrieved images for a particular query are marked with green border and falsely retrieved images are marked with red border.}
	\label{Image_ret}
\end{figure*}

\subsection{Impact of activation layer}
This experiment investigates the importance of applying activation functions to deep features. The baseline is the Direct aggregation (DA) method where the convolutional layer features are simply aggregated using average pooling. We compare the baseline performance with the proposed SinH aggregation (SinHA), Exponential aggregation (ExpA) and Weibull aggregation (WBA) methods. It can be observed from Figure \ref{RESULTS_ACT} that it is essential to apply activation functions to the features before aggregation to obtain significant increase in retrieval accuracy. The Weibull aggregation provides average gains of 10.5\%, 11.3\%, 4.9\% and 7.1\% on $R$Oxf$M$, $R$Oxf$H$, $R$Par$M$ and $R$Par$H$, versus Direct aggregation. Furthermore, the WBA network also outperforms ExpA and SinHA on all datasets. 

In all the following experiments and comparison with the state-of-the-art, we will use Weibull aggregation with ACTNET.

\subsection{Impact of multi-stream aggregation}
This section demonstrates the advantage of aggregating a hierarchy of convolutional features from multiple CNN layers. In ACTNET, deep features from the last two convolutional layers of ResNext101 are transformed using modified Weibull function, aggregated via average pooling and concatenated to form image representation. Importantly, the trainable parameters of the two aggregation streams are optimised explicitly to achieve significant retrieval improvement. It can be observed from Table \ref{Multimap} that multi-stream aggregation (MSA) brings an improvement of +1.0\%, +2.5\% and 3.1\% on Holidays, Oxford5k and MPEG datasets respectively compared to single stream aggregation (SSA), where only the last layer features are used to compute the global descriptor. The gain in retrieval accuracy is even more pronounced on difficult datasets: 6.7\% and 2.9\% on $R$Oxf$H$ and $R$Par$H$ datasets. 

\subsection{Comparison of ACTNET, GEM and REMAP networks} \label{sec:GEM,REMAP,ACTNET}
In this section we compare ACTNET with the modified GEM (multi-stream) and REMAP networks. For a fair assessment, we train ACTNET, GEM and REMAP in an end-to-end manner on GLRD dataset with triplet loss. We use multi-stream aggregation for all networks. 

In multi-stream GEM architecture, the convolutional features from both streams are aggregated using generalized mean pooling, concatenated, L2-normalised, PCA-Whitened and L2-normalised again to form image descriptor.

In REMAP architecture, the convolutional features from last two streams are passed through rigid grid ROI-pooling block. The region based descriptors are L2-normalized, scaled based on KL-divergence weights and sum-pooled. The two 2048 dimensional descriptors from each streams are concatenated, PCA-Whitened and L2-normalized to form REMAP signature. After training the networks, we extract $D$-dimensional ($D=4096$) ACTNET, GEM and REMAP representations from the test datasets and compute the retrieval performance (Figure \ref{REMAP_vs_RMAC}), as a function of descriptor dimensionality. For the comparison, we use single image scale ([1024 $\times$ 768]) to compute ACTNET, GEM and REMAP representations.

It can be observed from Figure \ref{REMAP_vs_RMAC} that ACTNET consistently achieves higher mAP than GEM and REMAP signatures on all datasets.

For qualitative understanding, Figure \ref{Image_ret} demonstrates two queries of $R$Oxf dataset where the difference in recall between ACTNET and GEM is most significant and one query where the difference in recall between GEM and ACTNET is the biggest. For every query, the correctly retrieved images are marked by a green frame.

\subsection{Multi-resolution representation}
The retrieval performance can be significantly improved by aggregating descriptors extracted from different input image resolutions. We follow \cite{Gordo2017} to compute descriptors from images resized at two scales, with 1448 and 1024 pixels in the larger side, preserving the aspect ratio. The two descriptors are sum-aggregated and L2-normalised to obtain multi-resolution ACTNET representation.

\begin{table}[h]
\caption{Comparison of single-resolution and multi-resolution ACTNET}
\centering
\label{MS_ACTNET}
\begin{tabular}{|c|c|c|c|c|c|c|}
\hline
& Oxford5k & $R$Oxf$M$ & $R$Oxf$H$ & $R$Par$M$ & $R$Par$H$ \\ \hline \hline
single-res & 91.4 & 72.7 & 52.2 & 80.7 & 62.7\\ \hline
multi-res  & 92.4 & 74.8 & 54.6 & 81.5 & 63.8 \\ \hline
\end{tabular}
\end{table}

It can be seen from Table \ref{MS_ACTNET} that multi-resolution representation improves mAP by 2.1\%, 2.4\% and 1.1\% on $R$Oxf$M$, $R$Oxf$H$ and $R$Par$H$ datasets compared to single-resolution ACTNET.

\subsection{Performance of low complexity ACTNET}

In this section, the retrieval performance of low depth, low latency and small memory Mobile-ACTNET is evaluated. The Mobile-ACTNET architecture consists of baseline MobileNetV2 followed by our multi-stream activation based aggregation system. The entire network is trained on GLRD dataset using triplet loss. To show the effectiveness of our method, we compare the performance of Mobile-ACTNET with the computationally expensive state-of-the-art ResNet101-GEM \cite{GEM} and ResNet101-RMAC \cite{Gordo2017}. We use single-scale representations of Mobile-ACTNET, ResNet101-GEM and ResNet101-RMAC. It is interesting to observe from Table \ref{MobileNettab} that Mobile-ACTNET achieves comparable performance than the aforementioned methods while having five times faster extraction speed. 

\begin{table}[h]
\caption{Mobile-ACTNET comparison with state-of-the-art ResNet101-GEM and ResNet101-RMAC representations}
\label{MobileNettab}
\begin{adjustbox}{width=\columnwidth}
\begin{tabular}{|l|c|c|c|c|c|}
\hline
 & Hol & Oxf5k & $R$Oxf$M$ & $R$Oxf$H$ & MPEG \\ \hline \hline
Mobile-ACTNET & 92.9 & 85.1 & 61.1 & 35.5 & 73.1 \\ \hline
ResNet101-GEM & 92.4 & 85.7 & 61.7 & 35.1 & 74.1 \\ \hline
ResNet101-RMAC & 94.0 & 84.1 & 60.9 & 32.4 & 72.2 \\ \hline
\end{tabular}
\end{adjustbox}
\end{table}

\begin{table*}[t]
\centering
\caption{Comparison with the state-of-the-art using full dimensional descriptor on Oxford5k, $R$Oxford Medium protocol ($R$Oxf$M$), $R$Oxford Hard protocol ($R$Oxf$H$), $R$Paris Medium protocol ($R$Par$M$), $R$Paris Hard protocol ($R$Par$H$), and MPEG without and with one Million distractors ($R$1M). All results are computed in terms of mAP(\%). The results for MPEG datasets are computed using the software provided by the authors. $^\dagger$ indicates multi-resolution methods.}
\label{full_SOTA}
\begin{tabular}{|l|c|c|c|c|c|c|c|c|c|c|c|}
\hline
 & Net & Oxford & $R$Oxf$M$ & $R$Oxf$H$ & $R$Oxf$M$+ & $R$Oxf$H$+ & $R$Par$M$ & $R$Par$H$ & $R$Par$M$+ & $R$Par$H$+ & MPEG \\ 
 &  &5k  & & &$R$1M    &  $R$1M   & & &  $R$1M   &  $R$1M        &     \\ \hline \hline
NetVLAD \cite{NetVLAD} & VGG & 71.6 & 37.1 & 13.8 & 20.7 & 6.0 & 59.8 & 35.0 & 31.8 & 11.5 & 65.1 \\ \hline
SPoC \cite{SPOC} & VGG & 68.1 & 39.8 & 12.4 & 21.5 & 2.8 & 69.2 & 44.7 & 41.6 & 15.3 & 63.2 \\ \hline
CroW \cite{Kalantidi}& VGG & 70.8 & 42.4 & 13.3 & 21.2 & 3.3 & 70.4 & 47.2 & 42.7 & 16.3 & 68.2 \\ \hline
MAC \cite{ROXF} & VGG & 80.0 & 41.7 & 18.0 & 24.2 & 5.7 & 66.2 & 44.1 & 40.8 & 18.2 & 70.6 \\ \hline
AGEM \cite{AGEM}& ResNet & - & 67.0 & 40.7 & - & - & 78.1 & 57.3 & - & - & - \\ \hline
FS.EGM \cite{FSGEM}& ResNet & - & 63.0 & 34.5 & - & - & 68.7 & 43.9 & - & - & - \\ \hline
OS.EGM \cite{FSGEM}& ResNet & - & 64.2 & 35.9 & - & - & 69.9 & 46.1 & -  & - & - \\ \hline
ASDA \cite{ASDA} & ResNet & 87.7 & 66.4 & 38.5 & - & - & 71.6 & 47.9 & -  & - & - \\ \hline
REMAP$^\dagger$ \cite{REMAP} & ResNext & 91.4 & 67.3 & 42.5 & 48.5 & 23.1 & 78.5 & 59.8 & 54.1 & 26.3 & 80.1 \\ \hline
GEM$^\dagger$ \cite{ROXF} & ResNet & 87.8 & 64.7 & 38.5 & 45.2 & 19.9 & 77.2 & 56.3 & 52.3 & 24.7  & 74.1 \\ \hline
RMAC$^\dagger$ \cite{ROXF}& ResNet& 86.1 & 60.9 & 32.4 & 39.3 & 12.5 & 78.9 & 59.4 & 54.8 & 28.0  & 72.2 \\ \hline
ResNext-GEM$^\dagger$ & ResNext & 88.3 & 66.5 & 41.8 & 45.3 & 20.2 & 79.7 & 60.4 & 54.9 & 26.8 & 77.1 \\ \hline
GEM (AP)$^\dagger$ \cite{GEMAP} & ResNet & - & 67.5 & 42.8 & 47.5 & 23.2 & 80.1 & 60.5 & 52.5 & 25.1 & - \\ \hline
ACTNET$^\dagger$ & ResNext & \textbf{92.4} & \textbf{74.8} & \textbf{54.6} & \textbf{54.9} & \textbf{29.9} & \textbf{81.5} & \textbf{63.8} & \textbf{56.8} & \textbf{28.9} & \textbf{82.9} \\ \hline
\end{tabular}
\end{table*}

\begin{table*}[]
\caption{Performance evaluation of full dimensional image signatures using Query Expansion.}
\centering
\label{QExp}
\begin{tabular}{|l|c|c|c|c|c|c|c|c|}
\hline
 & $R$Oxf$M$ & $R$Oxf$H$ & $R$Oxf$M$+ & $R$Oxf$H$+ & $R$Par$M$ & $R$Par$H$ & $R$Par$M$+ & $R$Par$H$+ \\ 
 & & &$R$1M &$R$1M & & &$R$1M&$R$1M \\ \hline \hline
DELF-D2R-R-ASMK+SP \cite{D2R}& 71.9 & 48.5 & - & - & 78.0 & 54.0 & - & - \\ \hline
DELF-GLD-ASMK+SP \cite{D2R} & 76.0 & 52.4 & - & - & 80.2 & 58.6 & - & - \\ \hline
GEM+$\alpha$QE \cite{ROXF}& 67.2 & 40.8 & 49.0 & 24.2 & 80.7 & 61.8 & 58.0 & 31.0 \\ \hline
RMAC+$\alpha$QE \cite{ROXF} & 64.8 & 36.8 & 45.7 & 19.5 & 82.7 & 65.7 & 61.0 & 35.0 \\ \hline
ResNext-GEM+$\alpha$QE & 71.3 & 45.5 & 52.7 & 26.1 & 82.7 & 65.5 & 62.5 & 36.8 \\ \hline
GEM (AP)+$\alpha$QE \cite{GEMAP} & 71.4 & 45.9 & 53.1 & 26.2 & 84.0 & 67.3 & 60.3 & 32.3 \\ \hline
ACTNET+$\alpha$QE & \textbf{79.7} & \textbf{59.7} & \textbf{63.2} & \textbf{37.3} & \textbf{84.7} & \textbf{68.7} & \textbf{66.5} & \textbf{42.1} \\ \hline \hline
GEM+DFS \cite{ROXF}& 69.8 & 40.5 & 61.5 & 33.1 & 88.9  & 78.5 & 84.9 & 71.6 \\ \hline 
RMAC+DFS \cite{ROXF} & 69.0 & 44.7 & 56.6 & 28.4 & 89.5  & 80.0 & 83.2 & 70.4 \\ \hline 
ACTNET+DFS & \textbf{79.3} & \textbf{59.3} & \textbf{70.5} & \textbf{49.1} & \textbf{91.7}  & \textbf{84.6} & \textbf{89.1} & \textbf{79.5} \\ \hline

\end{tabular}
\end{table*}

\begin{table*}[]
\caption{Comparison with the state-of-the-art using small dimensional descriptor (128-Dim). The results for RMAC and GEM are computed using the software provided by authors.}
\label{small_dim_SOTA}
\begin{tabular}{|l|c|c|c|c|c|c|c|c|c|c|c|}
\hline
 & Dim & Holidays & Oxford5k & $R$Oxf$M$ & $R$Oxf$H$ & $R$Oxf$M$+ & $R$Oxf$H$+ & $R$Par$M$ & $R$Par$H$ & $R$Par$M$+ & $R$Par$H$+ \\ 
 &  &   &  & &  & $R$1M    & $R$1M    &  &       & $R$1M       & $R$1M \\ \hline \hline
RMAC \cite{Gordo2017} & 128 & 88.5 & 77.9 & 46.2 & 21.1 & 21.9 & 7.4 & 72.3 & 48.1 & 40.9 & 15.6 \\ \hline 
GEM \cite{GEM}        & 128 & 85.9 & 79.5 & 49.9 & 26.2 & 27.8 & 10.9 & 72.2 & 47.5 & 40.8 & 15.5 \\ \hline
ACTNET  & 128 & \textbf{91.6} & \textbf{81.3} & \textbf{60.2} & \textbf{36.1} & \textbf{33.6} & \textbf{14.9} & \textbf{75.9} & \textbf{52.3} & \textbf{42.8} & \textbf{18.3} \\ \hline
\end{tabular}
\end{table*}

\section{Comparison with the state-of-the-art} \label{sec:SOTA}
We extensively compare the performance of ACTNET with the state-of-the-art on compact image signatures and on methods that use query expansion.

The performance of the full dimensional image signatures (1K-4K Dimensions) is summarised in Table \ref{full_SOTA}. In real world scenarios containing billions of images, the use of image representations with dimensionality greater than 1K is prohibitive due to memory requirements and search time; however the results are useful in understanding the maximum capabilities of each signatures. 

The proposed ACTNET consistently outperforms the state-of-the-art on all datasets. Compared to RMAC \cite{Gordo2017}, ACTNET provides a significant gain of +6.3\%, +13.9\%, 2.6\% and 10.6\% in mAP on Oxford5k, $R$Oxf$M$, $R$Par$M$ and MPEG datasets respectively. The difference in mAP points between ACTNET and RMAC is even higher on challenging datasets (+22.2\% $R$Oxf$H$, +15.6\% $R$Oxf$M$+$R$1M, +17.4\% $R$Oxf$H$+$R$1M, +4.4\% $R$Par$H$ and +2.0\% $R$Par$M$+$R$1M). Compared to the GEM representation \cite{GEM}, ACTNET is more than 16.1, 9.7, 10, 7.5 and 4.2 mAP points ahead on $R$Oxf$H$, $R$Oxf$M$+$R$1M, $R$Oxf$H$+$R$1M, $R$Par$H$ and $R$Par$H$+$R$1M datasets respectively. Our ACTNET outperforms most recent state-of-the-art method GEM (AP) \cite{GEMAP} on all datasets. We also compare ACTNET with our implementations of multi-resolution ResNext-GEM and REMAP and the retrieval performance shows that ACTNET creates significantly more robust and discriminative image signature.

It has recently become common practice to apply Query Expansion (QE) on top of global image signatures to further improve retrieval accuracy. More precisely, we use $\alpha$ query expansion \cite{GEM} and diffusion (DFS) \cite{iscen17cvpr}. It can be observed from Table \ref{QExp} that ACTNET+$\alpha$QE achieves significantly better mAP compared to RMAC+$\alpha$QE and GEM+$\alpha$QE. Furthermore, ACTNET+DFS achieves 70.5\%, 49.1\%, 89.1\% and 79.5\% on $R$Oxf$M$+$R$1M, $R$Oxf$H$+$R$1M, $R$Par$M$+$R$1M and $R$Par$H$+$R$1M datasets respectively, outperforming the best published results to date. It is also interesting to note that ACTNET+DFS outperforms the computationally complex DELF-D2R-R-ASMK+SP \cite{D2R} approach which uses CNN based local features, a codebook of 65k visual words and Spatial Verification (SP).

Finally, we compare the retrieval performance of compact (128 dimensional) image signatures which are practical in real world applications. To compute a compact signature, we forward pass an image through ACTNET to obtain 4096 dimensional representation. The top 128 values from 4096 forms the compact signature. The results presented in Table \ref{small_dim_SOTA} show that ACTNET significantly outperforms state-of-the-art RMAC and GEM methods. On large scale datasets of $R$Oxf$M$+$R$1M, $R$Oxf$H$+$R$1M, $R$Par$M$+$R$1M and $R$Par$H$+$R$1M, ACTNET achieves the best ever performances of 33.6\%, 14.9\%, 42.8\% and 18.3\% respectively.

\section{Conclusion} \label{sec:concl}
In this paper we introduced a novel CNN network architecture called ACTNET. The key innovation is a novel trainable activation layer designed to improve the signal-to-noise ratio in the aggregation stage of deep convolutional features. 
Our activation layer amplifies a few selected strong filter responses (corresponding to prominent features) thus reducing the impact of weaker features, which effectively constitute background noise. In ACTNET, deep features from final convolutional layers are transformed using the activation function and optimally aggregated into a global descriptor. 

The parameters of activation blocks are trained jointly with the CNN filter parameters in an end-to-end manner by minimising the triplet loss. We proposed and evaluated three parametric activation functions:  Sine-Hyperbolic, Exponential and modified Weibull, showing that while all bring significant gains, the Weibull achieved the best performance thanks to its equalising properties.

We provided a thorough evaluation on all key benchmarks demonstrating that ACTNET architecture with learnable aggregation generates global representations that significantly outperform the latest state-of-the-art methods, including those based on computationally costly local descriptor indexing and spatial verification. We also show that ACTNET retains its leading performance when using short codes of 128 bytes and when applied to low-complexity CNN.


%




\ifCLASSOPTIONcaptionsoff
  \newpage
\fi

\bibliographystyle{IEEEtran}
\bibliography{Activation_CNN}
\begin{IEEEbiography}[{\includegraphics[width=1in,height=1.2in]{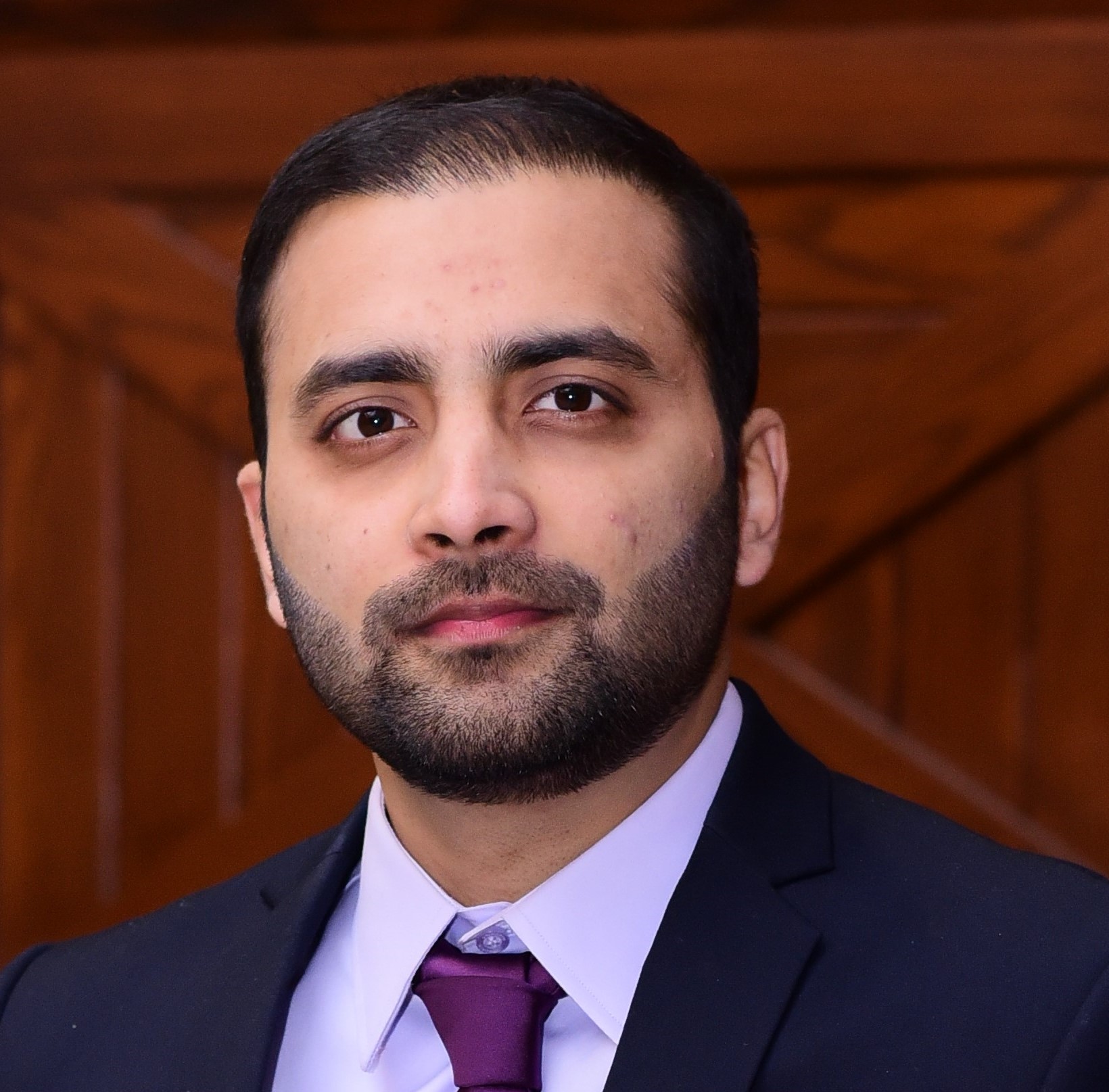}}]{Syed Sameed Husain} is a Research Fellow at the Centre for Vision, Speech and Signal Processing, University of Surrey, United Kingdom. He received MSc and PhD degrees from University of Surrey, in 2011 and 2016, respectively. His research interests include machine learning, computer vision, deep learning and image retrieval. Sameed's team has recently won the prestigious Google Landmark Retrieval Challenge. 
\end{IEEEbiography}

\begin{IEEEbiography}[{\includegraphics[width=1in,height=1.25in]{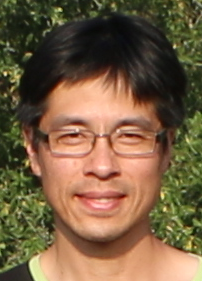}}]{Eng-Jon Ong} received the computer science degree in 1997 and the PhD degree in computer vision in 2001 from Queen Mary, University of London. Following that, he joined the Centre for Vision, Speech and Signal Processing at the University of Surrey as a researcher. His main interests are in visual feature tracking, data mining, pattern recognition, and machine learning methods.
\end{IEEEbiography}

\begin{IEEEbiography}[{\includegraphics[width=1in,height=1.25in]{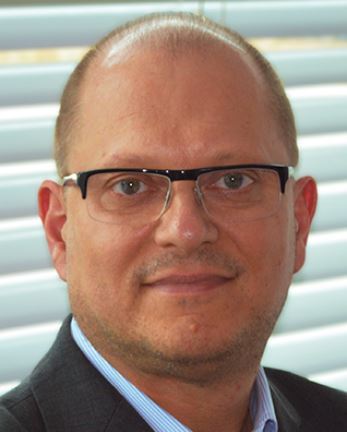}}]{Miroslaw Bober} is a Professor of Video Processing at the University of Surrey, U.K. In 2011 he co-founded Visual Atoms Ltd, a company specializing in visual analysis and search technologies. Between 1997 and 2011 he headed Mitsubishi Electric Corporate R\&D Center Europe (MERCE-UK). He received BSc degree from AGH University of Science and Technology, and MSc and PhD degrees from University of Surrey. His research interests include computer vision, machine learning and AI, with a focus on analysis and understanding of visual and multimodal data, and efficient representation of its semantic content. Miroslaw led the development of ISO MPEG standards for over 20 years, chairing the MPEG-7, CDVS and CVDA groups. He is an inventor of over 80 patents, many deployed in products. His publication record includes over 100 refereed publications, including three books and book chapters, and his visual search technologies recently won the Google Landmark Retrieval Challenge on Kaggle.
\end{IEEEbiography}


%

%

\end{document}